\documentclass[10pt,twocolumn,letterpaper]{article}

\usepackage{cvpr}
\usepackage{times}
\usepackage{epsfig}
\usepackage{graphicx}
\usepackage{amsmath}
\usepackage{amssymb}
\usepackage{booktabs}
\usepackage{bbm}
\usepackage{multirow}
\usepackage{subcaption}
\usepackage{makecell,rotating,multirow,diagbox}

\usepackage[pagebackref=true,breaklinks=true,letterpaper=true,colorlinks,bookmarks=false]{hyperref}

\cvprfinalcopy 

\def\cvprPaperID{****} 

\begin{document}

\title{Person Transfer GAN to Bridge Domain Gap for Person Re-Identification}

\author{Longhui Wei$^1$, Shiliang Zhang$^1$, Wen Gao$^1$, Qi Tian$^2$\\
\normalsize{$^1$Peking University}       \quad \normalsize{$^2$University of Texas at San Antonio}\\
\small{\{longhuiwei, slzhang.jdl, wgao\}@pku.edu.cn, qi.tian@utsa.edu}
}

\maketitle

\begin{abstract}
Although the performance of person Re-Identification (ReID) has been significantly boosted, many challenging issues in real scenarios have not been fully investigated, \emph{e.g.}, the complex scenes and lighting variations, viewpoint and pose changes, and the large number of identities in a camera network. To facilitate the research towards conquering those issues, this paper contributes a new dataset called MSMT17 with many important features, \emph{e.g.}, 1) the raw videos are taken by an 15-camera network deployed in both indoor and outdoor scenes, 2) the videos cover a long period of time and present complex lighting variations, and 3) it contains currently the largest number of annotated identities, \emph{i.e.}, 4,101 identities and 126,441 bounding boxes. We also observe that, domain gap commonly exists between datasets, which essentially causes severe performance drop when training and testing on different datasets. This results in that available training data cannot be effectively leveraged for new testing domains. To relieve the expensive costs of annotating new training samples, we propose a Person Transfer Generative Adversarial Network (PTGAN) to bridge the domain gap. Comprehensive experiments show that the domain gap could be substantially narrowed-down by the PTGAN.
\end{abstract}

\section{Introduction}

\begin{figure}
\begin{center}
\includegraphics[width=0.9\linewidth]{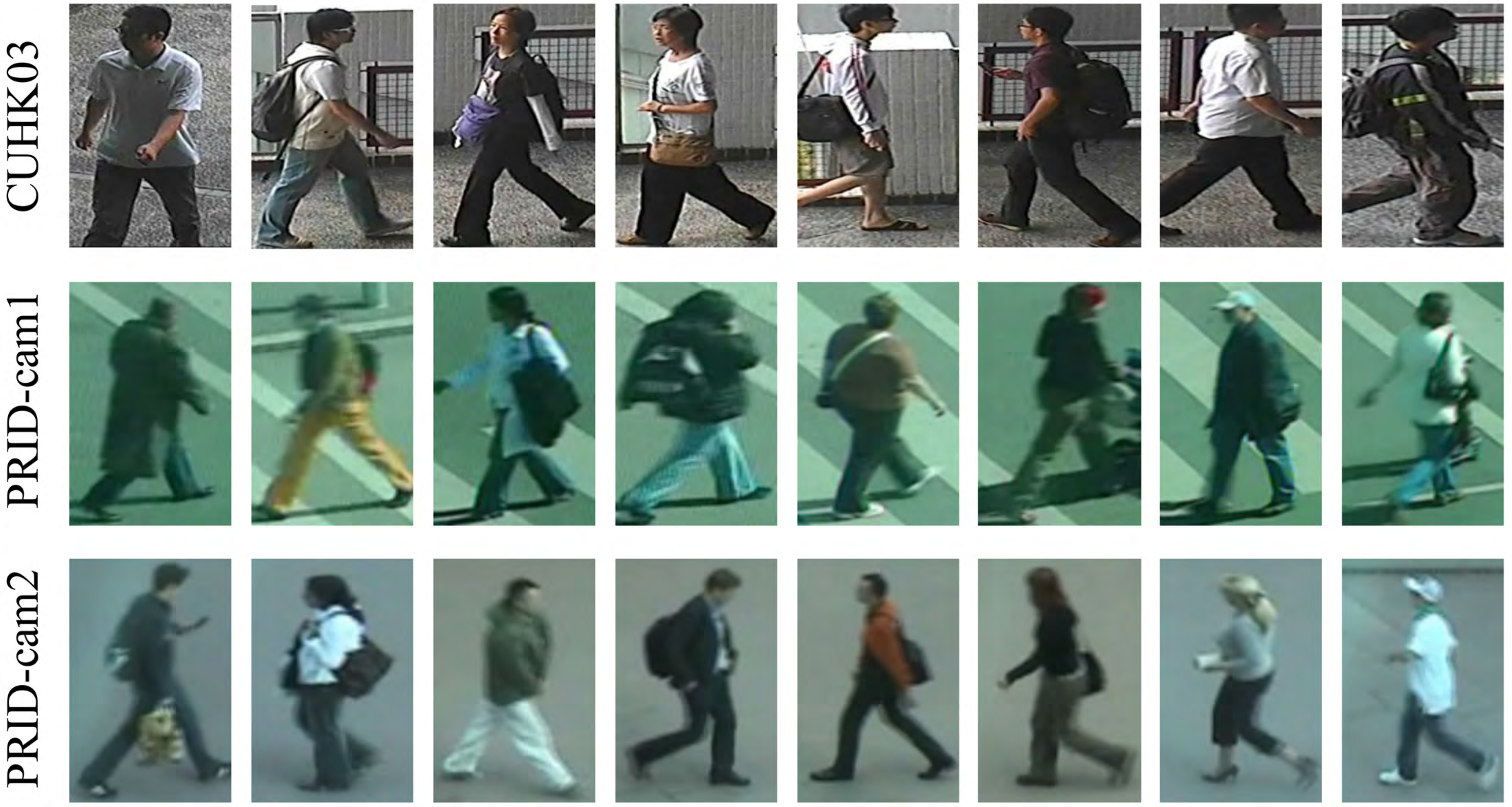}
\end{center}
  \vspace{-3mm}
\caption{Illustration of the domain gap between \emph{CUHK03} and \emph{PRID}. It is obvious that, \emph{CUHK03} and \emph{PRID} present different styles, \emph{e.g.}, distinct lightings, resolutions, human race, seasons, backgrounds, \emph{etc.}, resulting in low accuracy when training on \emph{CUHK03} and testing on \emph{PRID}.}
\label{fig:fig1}
\end{figure}

Person Re-Identification (ReID) targets to match and return images of a probe person from a large-scale gallery set collected by camera networks. Because of its important applications in security and surveillance, person ReID has been drawing lots of attention from both academia and industry. Thanks to the development of deep learning and the availability of many datasets, person ReID performance has been significantly boosted. For example, the Rank-1 accuracy of single query on \emph{Market1501}~\cite{zheng2015scalable} has been improved from 43.8\%~\cite{liao2015person} to 89.9\%~\cite{wei2017glad}. The Rank-1 accuracy on \emph{CUHK03}~\cite{li2014deepreid} labeled dataset has been improved from 19.9\%~\cite{li2014deepreid} to 88.5\%~\cite{su2017pose}. A more detailed review of current approaches will be given in Sec.~\ref{sec:relatedwork}.

Although the performance on current person ReID datasets is pleasing, there still remain several open issues hindering the applications of person ReID. First, existing public datasets differ from the data collected in real scenarios. For example, current datasets either contain limited number of identities or are taken under constrained environments. The currently largest \emph{DukeMTMC-reID}~\cite{zheng2017unlabeled} contains less than 2,000 identities and presents simple lighting conditions. Those limitations simplify the person ReID task and help to achieve high accuracy. In real scenarios, person ReID is commonly executed within a camera network deployed in both indoor and outdoor scenes and processes videos taken by a long period of time. Accordingly, real applications have to cope with challenges like a large number of identities and complex lighting and scene variations, which current algorithms might fail to address.

Another challenge we observe is that, there exists domain gap between different person ReID datasets, \emph{i.e.}, training and testing on different person ReID datasets results in severe performance drop. For example, the model trained on \emph{CUHK03}~\cite{li2014deepreid} only achieves the Rank-1 accuracy of 2.0\% when tested on \emph{PRID}~\cite{hirzer2011person}. As shown in Fig.~\ref{fig:fig1}, the domain gap could be caused by many reasons like different lighting conditions, resolutions, human race, seasons, backgrounds, \emph{etc}. This challenge also hinders the applications of person ReID, because available training samples cannot be effectively leveraged for new testing domains. Since annotating person ID labels is expensive, research efforts are desired to narrow-down or eliminate the domain gap.

Aiming to facilitate the research towards applications in realistic scenarios, we collect a new Multi-Scene Multi-Time person ReID dataset (\emph{MSMT17}). Different from existing datasets, \emph{MSMT17} is collected and annotated to present several new features. 1) The raw videos are taken by an 15-camera network deployed in both the indoor and outdoor scenes. Therefore, it presents complex scene transformations and backgrounds. 2) The videos cover a long period of time, \emph{e.g.}, four days in a month and three hours in the morning, noon, and afternoon, respectively in each day, thus present complex lighting variations. 3) It contains currently the largest number of annotated identities and bounding boxes, \emph{i.e.}, 4,101 identities and 126,441 bounding boxes. To our best knowledge, \emph{MSMT17} is currently the largest and most challenging public dataset for person ReID. More detailed descriptions will be given in Sec.~\ref{sec:pkudataset}.

To address the second challenge, we propose to bridge the domain gap by transferring persons in dataset \emph{A} to another dataset \emph{B}. The transferred persons from \emph{A} are desired to keep their identities, meanwhile present similar styles, \emph{e.g.}, backgrounds, lightings, \emph{etc.}, with persons in \emph{B}. We model this transfer procedure with a Person Transfer Generative Adversarial Network (PTGAN), which is inspired by the Cycle-GAN~\cite{zhu2017unpaired}. Different from Cycle-GAN~\cite{zhu2017unpaired}, PTGAN considers extra constraints on the person foregrounds to ensure the stability of their identities during transfer. Compared with Cycle-GAN, PTGAN generates high quality person images, where person identities are kept and the styles are effectively transformed. Extensive experimental results on several datasets show PTGAN effectively reduces the domain gap among datasets.

Our contributions can be summarized into three aspects. 1) A new challenging large-scale \emph{MSMT17} dataset is collected and will be released. Compared with existing datasets, \emph{MSMT17} defines more realistic and challenging person ReID tasks. 2) We propose person transfer to take advantages of existing labeled data from different datasets. It has potential to relieve the expensive data annotations on new datasets and make it easy to train person ReID systems in real scenarios. An effective PTGAN model is presented for person transfer. 3) This paper analyzes several issues hindering the applications of person ReID. The proposed \emph{MSMT17} and algorithms have potential to facilitate the future research on person ReID.

\section{Related Work} \label{sec:relatedwork}
This work is closely related with descriptor learning in person ReID and image-to-image translation by GAN. We briefly summarize those two categories of works in this section.

\subsection{Descriptor Learning in Person ReID}
Deep learning based descriptors have shown substantial advantages over hand-crafted features on most of person ReID datasets. Some works~\cite{xiao2016learning,zheng2017unlabeled} learn deep descriptors from the whole images with classification models, where each person ID is treated as a category. Some other works~\cite{zheng2016discriminatively,geng2016deep} combine verification models with classification models to learn descriptors. Hermans \emph{et al.}~\cite{hermans2017defense} show that triplet loss effectively improves the performance of person ReID. Similarly, Chen~\emph{et al.}~\cite{chen2017beyond} propose the quadruplet network to learn representations.

The above works learn global descriptors and ignore the detailed cues which might be important for distinguishing persons. To explicitly utilize local cues, Cheng~\emph{et al.}~\cite{cheng2016person} propose a multi-channel part-based network to learn a discriminative descriptor. Wu~\emph{et al.}~\cite{wu2016enhanced} discover hand-crafted features could be complementary with deep features. They divide the global image into five fixed-length regions. For each region, a histogram descriptor is extracted and concatenated with the global deep descriptor. Though the above works achieve good performance, they ignore the misalignment issue caused by fixed body part division. Targeting to solve this issue, Wei~\emph{et al.}~\cite{wei2017glad} utilize Deepercut~\cite{insafutdinov2016deepercut} to detect three coarse body regions and then learn an global-local-alignment descriptor. In~\cite{zhao2017spindle}, more fine-grained part regions are localized and then fed into the proposed Spindle Net for descriptor learning. Similarly, Li~\emph{et al.}~\cite{li2017learning} adopt Spatial Transform Networks (STN)~\cite{jaderberg2015spatial} to detect latent part regions and then learn descriptors on those regions.

\subsection{Image-to-Image Translation by GAN}
Since GAN proposed by Goodfellow \emph{et al.}~\cite{goodfellow2014generative}, many variants of GAN~\cite{mirza2014conditional,reed2016generative,wang2016generative,yoo2016pixel,li2016precomputed,yan2016attribute2image,ledig2016photo,gatys2016image,liu2016coupled,yi2017dualgan,kim2017learning,zhu2017unpaired} have been proposed to tackle different tasks,~\emph{e.g.}, natural style transfer, super-resolution, sketch-to-image generation, image-to-image translation, \emph{etc}. Among them, image-to-image translation has attracted lots of attention. In~\cite{isola2016image}, Isola \emph{et al.} propose conditional adversarial networks to learn the mapping function from input to output images. However, this method requires paired training data, which is hard to acquire in many tasks~\cite{zhu2017unpaired}. Targeting to solve the unpaired image-to-image translation task, Zhu~\emph{et al.}~\cite{zhu2017unpaired} propose cycle consistency loss to train unpaired data. Also, the works~\cite{yi2017dualgan,kim2017learning} propose a similar framework to solve the task. Our proposed PTGAN is similar to Cycle-GAN~\cite{zhu2017unpaired} in that, it also performs image-to-image translation. Differently, extra constraints on person identity are applied to ensure the transferred images can be used for model training. Zheng~\emph{et al.}~\cite{zheng2017unlabeled} adopt GAN to generate new samples for data augmentation in person ReID. Their work differs from ours in both motivation and methodology. As far as we know, this is an early work on person transfer by GAN for person ReID.

\section{\emph{MSMT17} Dataset}
\label{sec:pkudataset}

\begin{table*}
\footnotesize
\begin{center}
\caption{Comparison between \emph{MSMT17} and other person ReID datasets.}\label{tab:datasetCompare}
\resizebox{1.0\textwidth}{!}{%
\begin{tabular}{c|c|c|c|c|c|c|c|c}

\hline
{Dataset} &\textbf{\emph{MSMT17}}&{\emph{Duke}~\cite{zheng2017unlabeled,ristani2016MTMC}}&{\emph{Market}~\cite{zheng2015scalable}}&{\emph{CUHK03}~\cite{li2014deepreid}}&{\emph{CUHK01}~\cite{cuhk01}}&{\emph{VIPeR}~\cite{VIPeR}}&{\emph{PRID}~\cite{hirzer2011person}}&{\emph{CAVIAR}~\cite{cheng2011custom}}\\
\hline
\hline
{ BBoxes} &\textbf{126,441}&{36,411}&{32,668}&{28,192}&{3,884}&{1,264}&{1,134}&{610}\\
\hline
{ Identities} &\textbf{4,101}&{1,812}&{1,501}&{1,467}&{971}&{632}&{934}&{72}\\
\hline
{ Cameras} &\textbf{15}&{8}&{6}&{2}&{10}&{2}&{2}&{2}\\
\hline
{Detector} &\textbf{Faster RCNN}&{hand}&{DPM}&{DPM, hand}&{hand}&{hand}&{hand}&{hand}\\
\hline
{Scene} &\textbf{outdoor, indoor}&{outdoor}&{outdoor}&{indoor}&{indoor}&{outdoor}&{outdoor}&{indoor}\\
\hline
\end{tabular}}%
\end{center}
\end{table*}

\subsection{Overview of Previous Datasets}
Current person ReID datasets have significantly pushed forward the research on person ReID. As shown in Table~\ref{tab:datasetCompare}, \emph{DukeMTMC-reID}~\cite{zheng2017unlabeled}, \emph{CUHK03}~\cite{li2014deepreid}, and \emph{Market-1501}~\cite{zheng2015scalable} involve larger numbers of cameras and identities than \emph{VIPeR}~\cite{VIPeR} and \emph{PRID}~\cite{hirzer2011person}. The enough training data makes it possible to develop deep models and show their discriminative power in person ReID. Although current algorithms have achieved high accuracy on those datasets, person ReID is far from being solved and widely applied in real scenarios. Therefore, it is necessary to analyze the limitations of existing datasets.

Compared with the data collected in real scenarios, current datasets present limitations in four aspects: 1) The number of identities and cameras are not large enough, especially when compared with the real surveillance video data. In Table~\ref{tab:datasetCompare}, the largest dataset contains only 8 cameras and less than 2,000 identities. 2) Most of existing datasets cover only single scene, \emph{i.e.}, either indoor or outdoor scene. 3) Most of existing datasets are constructed from short-time surveillance videos without significant lighting changes. 4) Their bounding boxes are generated either by expensive hand drawing or out-dated detectors like Deformable Part Model (DPM)~\cite{felzenszwalb2010object}. Those limitations make it necessary to collect a larger and more realistic dataset for person ReID.

\subsection{Description to \textbf{\emph{MSMT17}}}
\label{sec:description}

\begin{figure}[ht!]
    \begin{center}
    \includegraphics[width=1\linewidth]{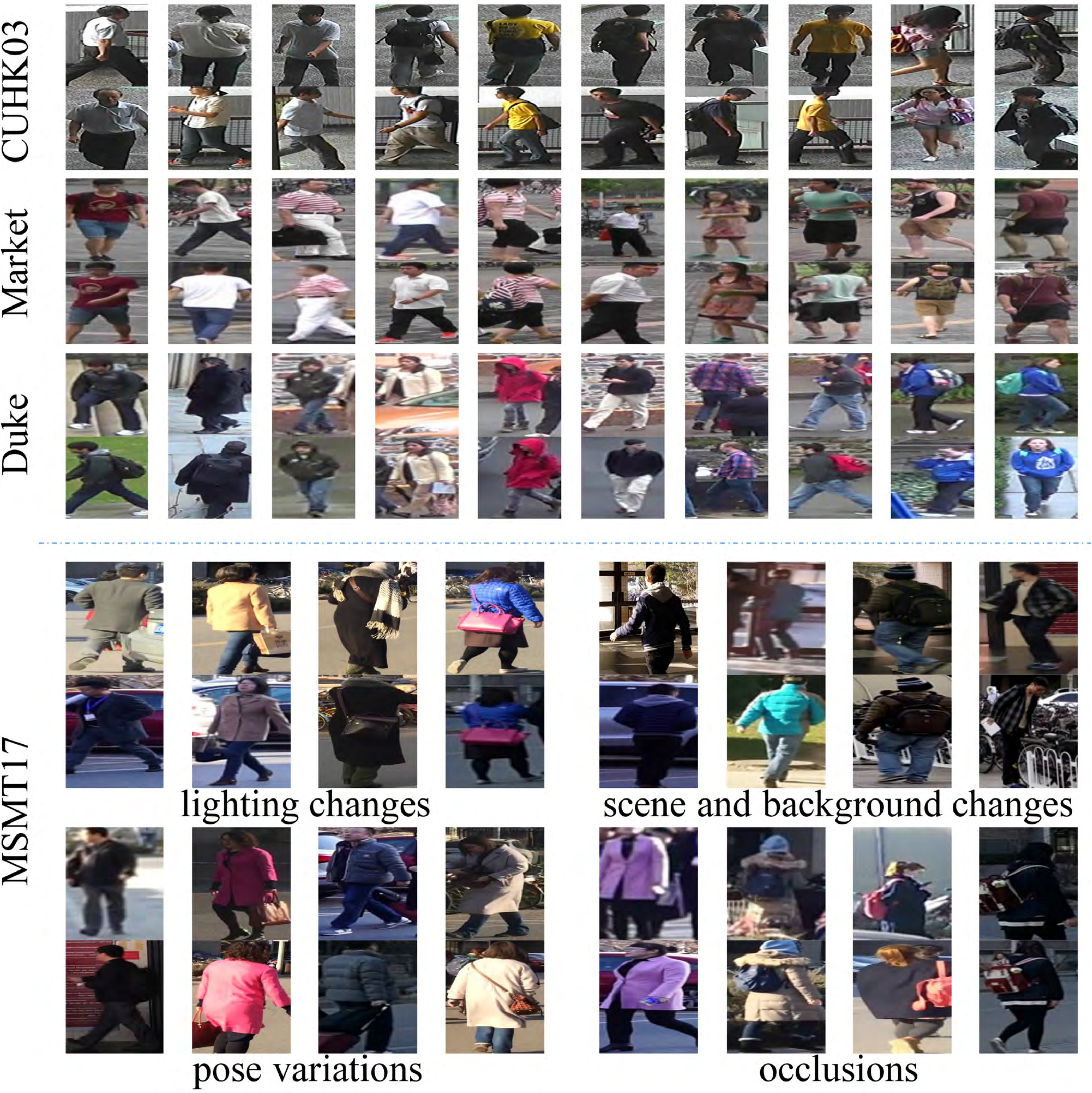}
    \end{center}
    \caption{Comparison of person images in \emph{CUHK03}, \emph{Market1501}, \emph{DukeMTMC-reID}, and \emph{MSMT17}. Each column shows two sample images of the same identity. It is obvious that, \emph{MSMT17} presents a more challenging and realistic person ReID task.}
    \label{fig:datasetcompare}
\end{figure}

\begin{figure*}[ht!]
    \begin{center}
    \includegraphics[width=1\linewidth]{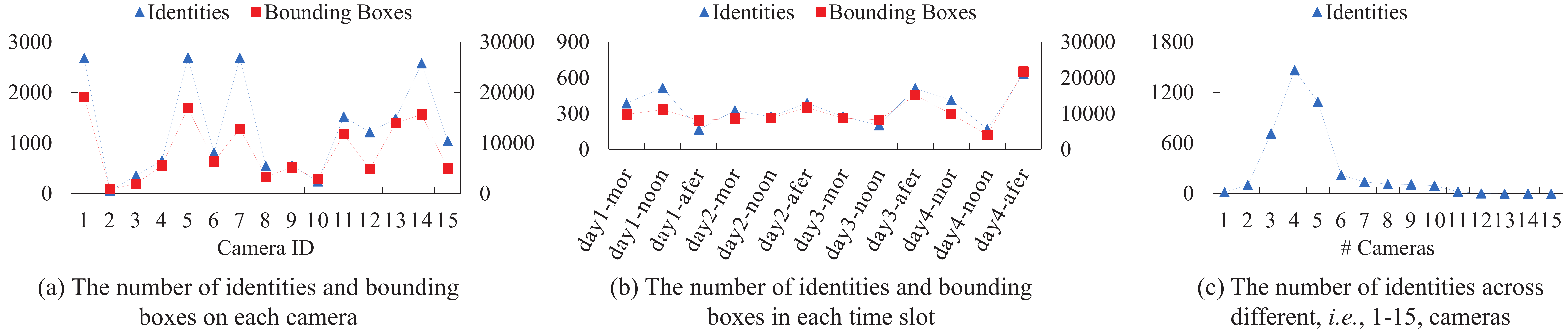}
    \end{center}
    \vspace{-3mm}
    \caption{Statistics of \emph{MSMT17}. }
    \label{fig:statistics}
\end{figure*}

Targeting to address above mentioned limitations, we collect a new Multi-Scene Multi-Time person ReID dataset (\emph{MSMT17}) by simulating the real scenarios as much as possible. We utilize an 15-camera network deployed in campus. This camera network contains 12 outdoor cameras and 3 indoor cameras. We select 4 days with different weather conditions in a month for video collection. For each day, 3 hours of videos taken in the morning, noon, and afternoon, respectively, are selected for pedestrian detection and annotation. Our final raw video set contains 180 hours of videos, 12 outdoor cameras, 3 indoor cameras, and 12 time slots. Faster RCNN~\cite{ren2015faster} is utilized for pedestrian bounding box detection. Three labelers go through the detected bounding boxes and annotate ID label for 2 months. Finally, 126,441 bounding boxes of 4,101 identities are annotated. Some statistics on \emph{MSMT17} are shown in Fig.~\ref{fig:statistics}. Sample images from \emph{MSMT17} are shown and compared in Fig.~\ref{fig:datasetcompare}. Compared with existing datasets, we summarize the new features in \emph{MSMT17} into the following aspects:

1) \emph{Larger number of identities, bounding boxes, and cameras}. To our best knowledge, \emph{MSMT17} is currently the largest person ReID dataset. As shown by the comparison in Table~\ref{tab:datasetCompare}, \emph{MSMT17} contains 126,441 bounding boxes, 4,101 identities, which are significantly larger than the ones in previous datasets.

2) \emph{Complex scenes and backgrounds}. \emph{MSMT17} contains the largest number of cameras, \emph{i.e.}, 15 cameras placed in different locations. It is also constructed with both indoor and outdoor videos, which has not been considered in previous datasets. Those considerations result in complex backgrounds and scene variations, also make \emph{MSMT17} more appealing and challenging.

3) \emph{Multiple time slots result in severe lighting changes}. \emph{MSMT17} is collected with 12 time slots, \emph{i.e.}, morning, noon, and afternoon in four days. It better simulates the real scenarios than previous datasets, but brings severe lighting changes.

4) \emph{More reliable bounding box detector}. Compared with hand drawing and DPM detector, Faster RCNN~\cite{ren2015faster} is a better choice for bounding box detection in real applications, \emph{e.g.}, easier to implement and more accurate.

\subsection{Evaluation Protocol}
\label{sec:evaluationProtocol}

We randomly divide our dataset into training set and testing set, respectively. Different from dividing the two parts equally in previous datasets, we set the training and testing ratio as 1:3. We use this setting because of the expensive data annotation in real scenarios, and thus want to encourage more efficient training strategies. Finally, the training set contains 32,621 bounding boxes of 1,041 identities, and the testing set contains 93,820 bounding boxes of 3,060 identities. From the testing set, 11,659 bounding boxes are randomly selected as query images and the other 82,161 bounding boxes are used as gallery images.

Similar with most of previous datasets, we utilize the Cumulated Matching Characteristics (CMC) curve to evaluate the ReID performance. For each query bounding box, multiple true positives could be returned. Therefore, we also regard person ReID as a retrieval task. mean Average Precision (mAP) is thus also used as the evaluation metric.

\section{Person Transfer GAN}

To better leverage the training set of dataset $A$ in person ReID tasks on dataset $B$, we propose to bridge the domain gap by transferring persons in \emph{A} to \emph{B}. As illustrated in Fig.~\ref{fig:fig1}, different datasets present distinct styles due to multiple reasons such as backgrounds, lighting conditions, resolutions, \emph{etc}. Imagine that, if persons in \emph{A} were captured by the cameras of \emph{B}, the style of those person images would be consistent with the style of \emph{B}. Our person transfer tries to simulate this procedure, \emph{i.e.}, learning a transfer function to 1) ensure the transferred person images show similar styles with the target dataset, and 2) keep the appearance and identity cues of the person during transfer.

This transfer task seems easy, \emph{e.g.}, can be finished by cropping the person foregrounds from \emph{A} and paste them on the backgrounds on \emph{B}. However, it is difficult to deal with multiple reasons of domain gap in a rule-based algorithm. Moreover, there could be complicated style variations on \emph{B}, \emph{e.g}., different backgrounds and lighting conditions between two cameras of \emph{PRID} in Fig.~\ref{fig:fig1}. Our algorithm is inspired by the popularity of GAN models, which have been proven effective in generating the desired image samples. We hence design a Person Transfer GAN (PTGAN) to perform person transfer from \emph{A} to \emph{B}.

Based on the above discussions, PTGAN is constructed to satisfy two constraints, \emph{i.e.}, the style transfer and person identity keeping. The goal of style transfer is to learn the style mapping functions between different person datasets. The goal of person identity keeping is to ensure the identity of one person remains unchanged after transfer. Because different transferred samples of one person are regarded as having the same person ID, the constraint on person identity is important for person ReID training. We thus formulate the loss function of PTGAN as, \emph{i.e.},
\begin{equation}
\begin{split}
\mathcal{L}_{PTGAN} = &\mathcal{L}_{Style} + \lambda_{1}\mathcal{L}_{ID},
\end{split}
\end{equation}
where $\mathcal{L}_{Style}$ denotes the style loss and $\mathcal{L}_{ID}$ denotes the identity loss, and $\lambda_{1}$ is the parameter for the trade-off between two losses.

ReID datasets do not contain paired person images, \emph{i.e.}, images of the same person from different datasets. Therefore, the style transfer can be regarded as an unpaired image-to-image translation task. Because of the good performance of Cycle-GAN in unpaired image-to-image translation task, we employ Cycle-GAN to learn the style mapping functions between dataset \emph{A} and \emph{B}. Suppose $G$ represents the style mapping function from \emph{A} to \emph{B} and $\overline{G}$ represents the style mapping function from \emph{B} to \emph{A}. $D_A$ and $D_B$ are the style discriminators for \emph{A} and \emph{B}, respectively. The objective function of style transfer learning can be formulated as follows:
\begin{equation}
\begin{split}
\mathcal{L}_{Style} =  &\mathcal{L}_{GAN}(G,D_B,A,B) \\
&+ \mathcal{L}_{GAN}(\overline{G},D_A,B,A) \\
&+ \lambda_{2} \mathcal{L}_{cyc}(G,\overline{G}),
\end{split}
\end{equation}
Where $\mathcal{L}_{GAN}$ represents the standard adversarial loss~\cite{goodfellow2014generative}, and $\mathcal{L}_{cyc}$ represents the cycle consistency loss~\cite{zhu2017unpaired}. For more details of those loss functions, please refer to the Cycle-GAN~\cite{zhu2017unpaired}.

Solely considering style transfer may result in ambiguous person ID labels in transferred person images. We thus compute the identity loss to ensure the accuracy of person ID labels in the transferred data. The person identity loss is computed by first acquiring the foreground mask of a person, then evaluating the variations on the person foreground before and after person transfer. Given the data distribution of \emph{A} as $a \sim p_{data}(a)$ and the data distribution of \emph{B} as $b \sim p_{data}(b)$. The objective function of identity loss can be formulated as follows:
\begin{equation}
\begin{split}
\mathcal{L}_{ID} = &\mathbbm{E}_{a \sim p_{data}(a)}[||(G(a)-a) \odot M(a)||_2]\\
&+ \mathbbm{E}_{b \sim p_{data}(b)}[||(\overline{G}(b)-b) \odot M(b)||_2],
\end{split}
\end{equation}
where $G(a)$ represents the transferred person image from image $a$, and $M(a)$ represents the foreground mask of person image $a$.

Because of its good performance on segmentation task, we use PSPNet~\cite{zhao2017pspnet} to extract the mask on person images. On video surveillance data with moving foregrounds and fixed backgrounds, more accurate and efficient foreground extraction algorithms can be applied. It can be observed that, PTGAN does not require person identity labels on the target dataset \emph{B}. The style discriminator $D_B$ can be trained with unlabled person images on \emph{B}. Therefore, PTGAN is well-suited to real scenarios, where the new testing domains have limited or no labeled training data.

\begin{figure}
\begin{center}
\includegraphics[width=1\linewidth]{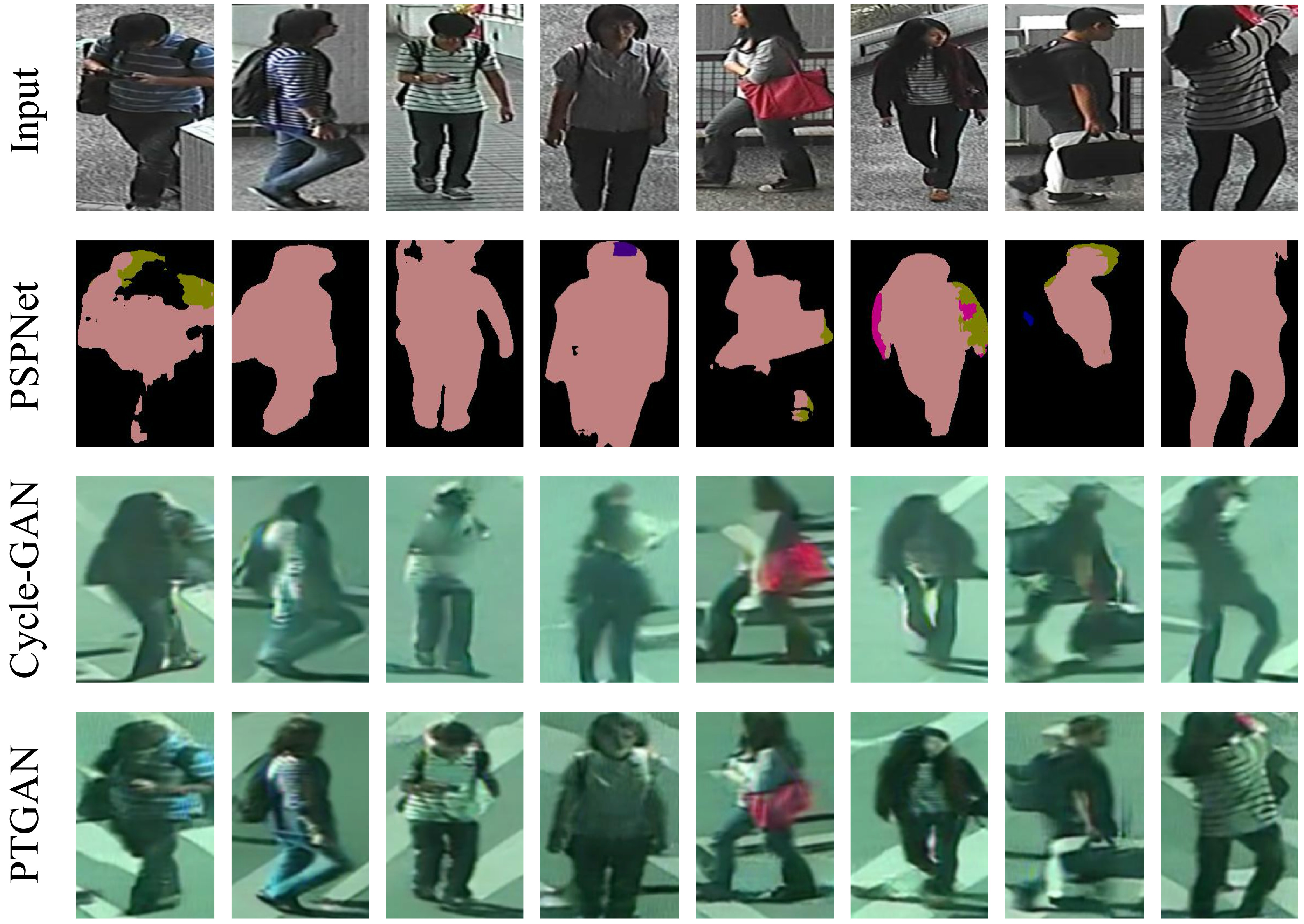}
\end{center}
\caption{Comparison of the transferred images by PTGAN and Cycle-GAN from \emph{CUHK03} to \emph{PRID-cam1}. The second row shows the segmentation results by PSPNet. The pink regions are segmented as person body regions.}
\label{fig:fig3}
\end{figure}

We show some sample results generated by PTGAN in Fig.~\ref{fig:fig3}. Compared with Cycle-GAN, PTGAN generates images with substantially higher quality. For example, the appearance of person is maintained and the style is effectively transferred toward the one on \emph{PRID} camera1. The shadows, road marks, and backgrounds are automatically generated and are similar with the ones on \emph{PRID} camera1. It is also interesting to observe that, PTGAN still works well with the noisy segmentation results generated by PSPNet. This implies that, PTGAN is also robust to the segmentation errors. More detailed evaluation of PTGAN will be given in Sec.~\ref{sec:exp_ptgan}.

\section{Experiments}

\subsection{Datasets}
In addition to the \emph{MSMT17}, four widely used person ReID datasets are employed in our experiments.

\emph{DukeMTMC-reID}~\cite{zheng2017unlabeled} is composed of 1,812 identities and 36,411 bounding boxes. 16,522 bounding boxes of 702 identities are used for training. The rest identities are included in the testing set. \emph{DukeMTMC-reID} is also denoted as \emph{Duke} for short.

\emph{Market-1501}~\cite{zheng2015scalable} contains 1,501 identities and 32,668 bounding boxes. The training set contains 12,936 bounding boxes of 751 identities. The rest 750 identities are included in the testing set. \emph{Market-1501} is also denoted as \emph{Market} for short.

\emph{CUHK03}~\cite{li2014deepreid} consists of 1,467 identities and 28,192 bounding boxes generated by both DPM and hand. Following the work~\cite{xiao2016learning},  26,264 bounding boxes of 1,367 identities are used for training, and 1,928 bounding boxes of 100 identities are used for testing.

\emph{PRID}~\cite{hirzer2011person} is composed of 934 identities from two cameras. Our experiments use the bounding boxes of 200 persons shared by both cameras as testing set.

\subsection{Implementation Details}
PTGAN uses similar network architecture with the one in Cycle-GAN~\cite{zhu2017unpaired}. For the generator network, two stride-$2$ convolutions, $9$ residual blocks, and two stride-$\frac{1}{2}$ fractionally-strided convolutions are designed. Two parts are included in the discriminator network. PatchGAN~\cite{isola2016image} is adopted as one part. The PatchGAN classifies whether a $70 \times 70$ patch in an image is real or fake. For the other part, $L_2$ distance between the transferred image and input image is computed on the foreground person.

Adam solver~\cite{kinga2015method} is adopted in PTGAN. For the generator network, the learning rate is set as 0.0002. The learning rate is set as 0.0001 for the discriminator network. We set $\lambda_{1} = 10$, and $\lambda_{2}=10$. The size of input image is $256 \times 256$. Finally, we train PTGAN for 40 epochs.

\subsection{Performance on \textbf{\emph{MSMT17}}}
\label{sec:MSMTTest}
As described in Sec.~\ref{sec:pkudataset}, \emph{MSMT17} is challenging but close to the reality. This section verifies this claim by testing existing algorithms on \emph{MSMT17}.

We go through the state-of-the-art works published in 2017 and 2016. Among those works, the GLAD proposed by Wei~\emph{et al.}~\cite{wei2017glad} achieves the best performance on \emph{Market}, and the PDC proposed by Su~\emph{et al.}~\cite{su2017pose} achieves the best performance on \emph{CUHK03}.\footnote{The work~\cite{liu2017hydraplus} reports better performance, but it is trained on an augmented data including training sets from three datasets.} We thus evaluate those two methods on \emph{MSMT17} with the codes and models provided by their authors. In most of person ReID works, GoogLeNet~\cite{szegedy2015going} is commonly used as the baseline model. We thus also use GoogLeNet~\cite{szegedy2015going} as our baseline.

\captionsetup{belowskip=-0cm}
\captionsetup{aboveskip=0.1cm}
\begin{table}
\footnotesize
\begin{center}
\caption{The performance of the state-of-the-art methods on \emph{MSMT17}. R-1 represents the Rank-1 accuracy.}\label{tab:tab4}
\begin{tabular}{c|c|c|c|c|c}
\hline
{Methods} &mAP&R-1 &R-5 &R-10 &R-20\\
\hline
\hline
GoogLeNet~\cite{szegedy2015going}     &23.0&47.6 &65.0&71.8&78.2\\
\hline
PDC~\cite{su2017pose} 	&29.7&58.0 &73.6 &79.4 &84.5\\
\hline
GLAD~\cite{wei2017glad}      &\textbf{34.0} &\textbf{61.4}&\textbf{76.8}&\textbf{81.6}&\textbf{85.9}\\
\hline
\end{tabular}
\end{center}
\end{table}

We summarize the experimental results in Table~\ref{tab:tab4}. As shown in the table, the baseline only achieves mAP of 23\% on \emph{MSMT17}, which is significantly lower than its mAP of 51.7\% on \emph{Market}~\cite{geng2016deep}. It is also obvious that, PDC~\cite{su2017pose} and GLAD~\cite{wei2017glad} substantially outperform the baseline performance by considering extra part and regional features. However, the best performance achieved by GLAD, \emph{e.g.}, mAP of 34\%, is still substantially lower than its reported performance on other datasets, \emph{e.g.}, 73.9\% on \emph{Market}. The above experiments clearly show the challenges of \emph{MSMT17}.

\begin{figure}
\begin{center}
\includegraphics[width=1\linewidth]{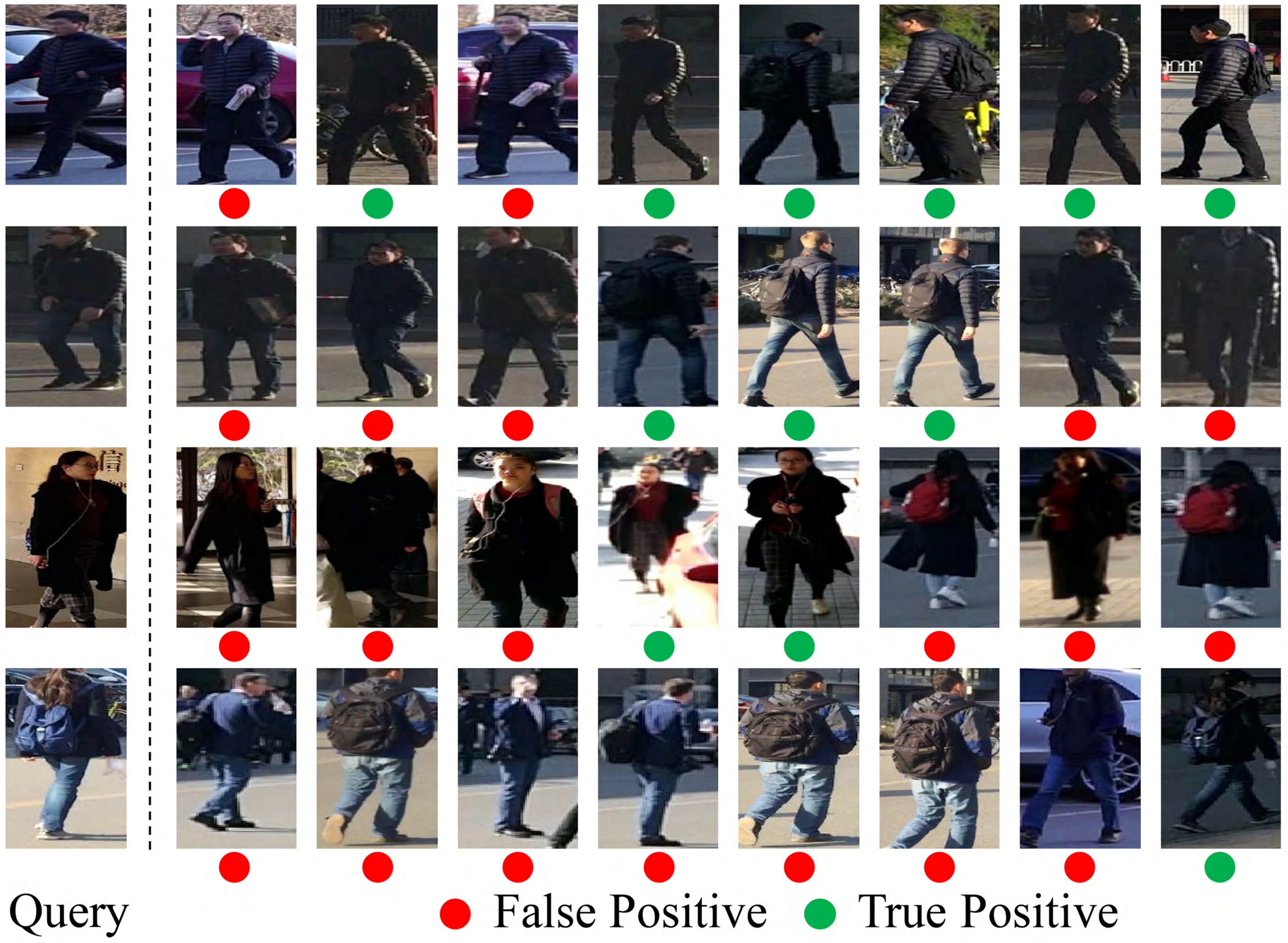}
\end{center}
\caption{Sample person ReID results generated by the method of GLAD~\cite{wei2017glad} on \emph{MSMT17}.}
\label{fig:ranklist}
\end{figure}

We also show some sample retrieval results in Fig.~\ref{fig:ranklist}. From the samples, we can conclude that although challenging, the ReID task defined by \emph{MSMT17} is realistic. Note that, in real scenarios distinct persons may present similar clothing cues, and images of same person may present different lightings, backgrounds, and poses. As shown in Fig.~\ref{fig:ranklist}, the false positive samples do show similar appearances with the one of query person. Some true positives present distinct lightings, poses, and backgrounds from the query. Therefore, we believe \emph{MSMT17} is a valuable dataset to facilitate the future research on person ReID.

\subsection{Performance of Person Transfer}  \label{sec:exp_ptgan}
Person transfer is performed from dataset \emph{A} to \emph{B}. The transferred data is hence used for training on \emph{B}. To ensure there is enough transferred data for training on \emph{B}, we test person transfer in two cases, \emph{i.e.}, 1) transferring from a large $A$ to a small $B$, and 2) transferring from a large $A$ to a large $B$. In the following experiments, we use the training set provided by \emph{A} for person transfer.

\subsubsection{Transfer from Large Dataset to Small Dataset}
\label{sec:BigToSmall}

This part tests the performance of transferred person data from \emph{CUHK03} and \emph{Martket} to a small dataset \emph{PRID}. As shown in Fig.~\ref{fig:fig1}, person images captured by two cameras on \emph{PRID} show different styles. Therefore, we perform person transfer to those two cameras, \emph{i.e.}, \emph{PRID-cam1} and \emph{PRID-cam2}, respectively.

\begin{figure}
\begin{center}
\includegraphics[width=1\linewidth]{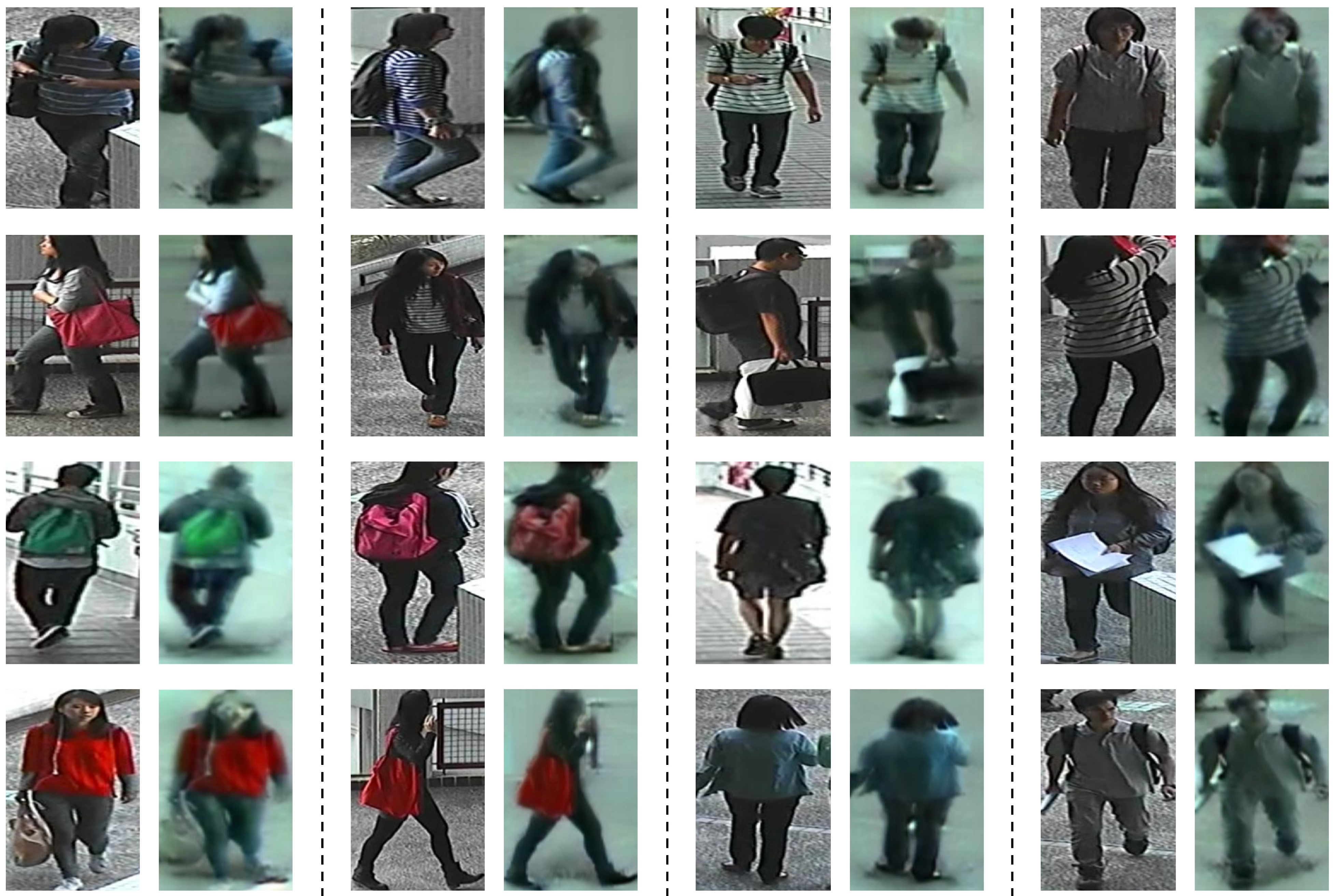}
\end{center}
\caption{Sample transferred person images from \emph{CUHK03} to \emph{PRID-cam2}. Each sample shows an image from \emph{CUHK03} in the first column, and the transferred image in the second column. }
\label{fig:fig5}
\end{figure}

\begin{figure}
\begin{center}
\includegraphics[width=1\linewidth]{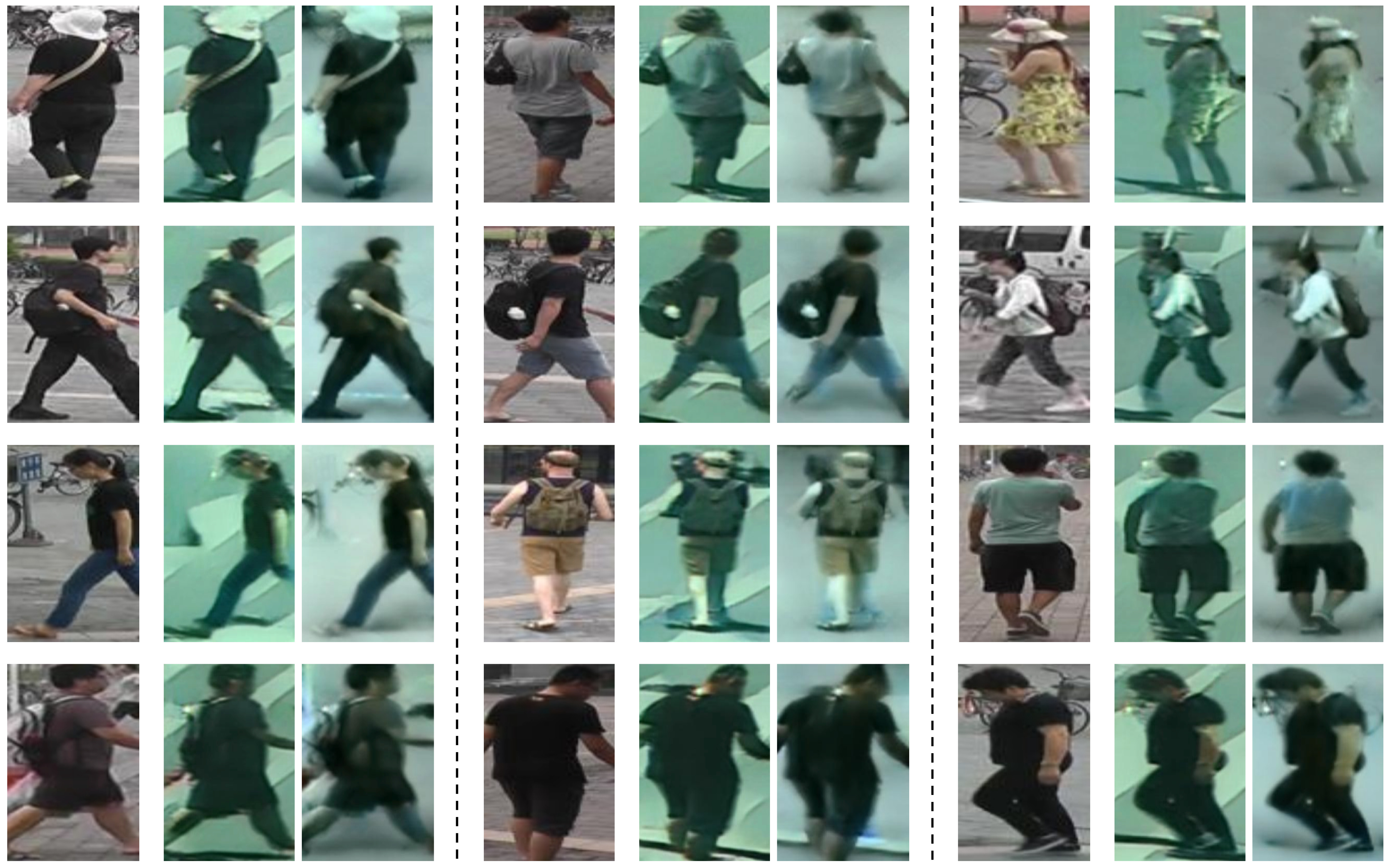}
\end{center}
\caption{Sample transferred person images from \emph{Market} to \emph{PRID-cam1} and \emph{PRID-cam2}. Images in the first column are from \emph{Market}. Transferred images to \emph{PRID-cam1} and \emph{PRID-cam2} are shown in the second and third columns, respectively.}
\label{fig:fig6}
\end{figure}

We first perform person transfer from \emph{CUHK03} to \emph{PRID-cam1} and \emph{PRID-cam2}. Samples of the transferred person images to \emph{PRID-cam1} are shown in Fig.~\ref{fig:fig3}. We additionally show samples of transferred person images from \emph{CUHK03} to \emph{PRID-cam2} in Fig.~\ref{fig:fig5}. It is clear that, the transferred person images to those two cameras show different styles, which are consistent with the ones on \emph{PRID}. We also transfer \emph{Market} to \emph{PRID-cam1} and \emph{PRID-cam2}, respectively. Samples of the transferred person images from \emph{Market} are shown in Fig.~\ref{fig:fig6}, where similar results can be observed as the ones in Fig.~\ref{fig:fig3} and Fig.~\ref{fig:fig5}, respectively.

\captionsetup{belowskip=-0cm}
\captionsetup{aboveskip=0.1cm}
\begin{table}

\footnotesize
\begin{center}
\caption{Performance of GoogLeNet tested on \emph{PRID} but trained with different training sets. $*$ denotes the transferred dataset. For instance, the subscript ${cam1}$ represents the transferred target dataset \emph{PRID-cam1}. ``cam1/cam2" means using images in \emph{PRID-cam1} as query set and images from \emph{PRID-cam2} as gallery set. }\label{tab:tab2}
\begin{tabular}{c|c|c|c|c}
\hline
\multirow{2}{*}{Training Set}       &\multicolumn{2}{c|}{cam1/cam2}&\multicolumn{2}{c}{cam2/cam1} \\

 \cline{2-5} &R-1&R-10        &R-1&R-10\\

\hline
\hline

\emph{CUHK03}     &2.0&11.5 &1.5&11.5\\
\emph{CUHK03}$^{*}_{cam1}$       &18.0&43.5 &6.5&24.0\\
\emph{CUHK03}$^{*}_{cam2}$        &17.5&53.0 &22.5&54.0\\
\emph{CUHK03}$^{*}_{cam1}$  +  CUHK03$^{*}_{cam2}$       &\textbf{37.5}&\textbf{72.5} &\textbf{37.5}&\textbf{69.5}\\
\hline

\emph{Market}     &5.0&26.0 &11.0&40.0\\
\emph{Market}$^{*}_{cam1}$        &17.5&50.5 &8.5&28.5\\
\emph{Market}$^{*}_{cam2}$       &10.0&31.5 &10.5&37.5\\
\emph{Market}$^{*}_{cam1}$  +  \emph{Market}$^{*}_{cam2}$       &\textbf{33.5}&\textbf{71.5} &\textbf{31.0}&\textbf{70.5}\\
\hline
\end{tabular}
\end{center}
\end{table}

To further evaluate whether the domain gap is reduced through PTGAN. We conduct comparisons between GoogLeNet trained with the training sets on \emph{CUHK03} and \emph{Market}, and GoogLeNet trained on their transferred training sets, respectively. The experimental results are summarized in Table~\ref{tab:tab2}. As shown in the table, GoogLeNet trained on the \emph{CUHK03}, only achieves the Rank-1 accuracy of 2.0\% on \emph{PRID}, which implies substantial domain gap between \emph{CUHK03} and \emph{PRID}. With training data transferred by PTGAN, GoogLeNet achieves a significant performance boost, \emph{e.g.}, the Rank-1 accuracy is improved from \textbf{2.0\%} to \textbf{37.5\%}, the Rank-10 accuracy is improved from \textbf{11.5\%} to \textbf{72.5\%}. Similar improvements can be observed from the results on \emph{Martket}, \emph{e.g.}, the Rank-1 accuracy is significantly improved from 5.0\% to 33.5\% after person transfer. The substantial performance improvements clearly indicate the shrunken domain gap. Moreover, this experiment shows that even without using labeled data on \emph{PRID}, we can achieve reasonable performance on it using training data from other datasets.

From Table~\ref{tab:tab2}, we also observe an interesting phenomenon,\emph{ i.e.}, combining the transferred datasets on two cameras results in better performance. This might be due to two reasons: 1) the combined dataset has more training samples, thus helps to train a better deep network, and 2) it enables the learning of style differences between two cameras. In the combined dataset, each person image has two transferred samples on camera1 and camera2, respectively with different styles. Because those two samples have the same person ID label, this training data enforces the network learning to gain robustness to the style variations between camera1 and camera2.

\subsubsection{Transfer from Large Dataset to Large Dataset}
\label{sec:BigToBig}

\begin{figure*}[ht!]
    \begin{center}
    \includegraphics[width=0.95\linewidth]{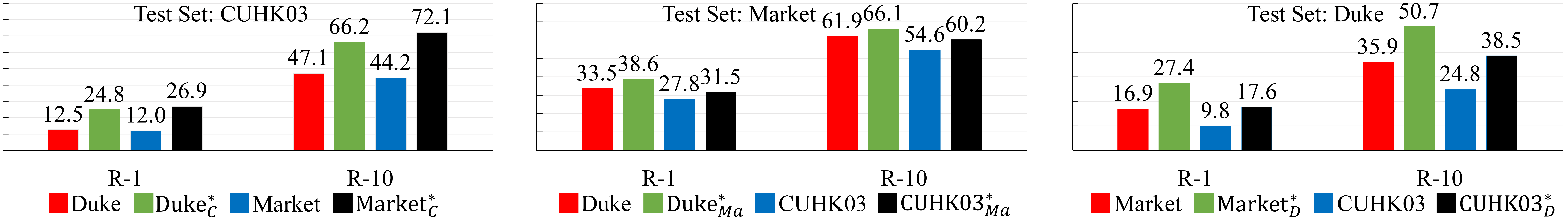}
    \end{center}
    \caption{Rank-1 and Rank-10 accuracies of GoogLeNet on \emph{CUHK03}, \emph{Market}, and \emph{Duke}. The subscripts ${C}$, ${Ma}$, and ${D}$ denote the transferred target dataset is \emph{CUHK03}, \emph{Market}, and \emph{Duke}, respectively. }
    \label{fig:PTGAN_comp}
\end{figure*}

This part simulates a more challenging scenario commonly existing in real applications, \emph{i.e.}, the available training data on a large testing set is not provided. We thus test the performance of PTGAN by conducting person transfer among three large datasets, \emph{i.e.}, \emph{Duke}, \emph{Market}, and \emph{CUHK03}, respectively.

\begin{figure}
\begin{center}
\includegraphics[width=0.9\linewidth]{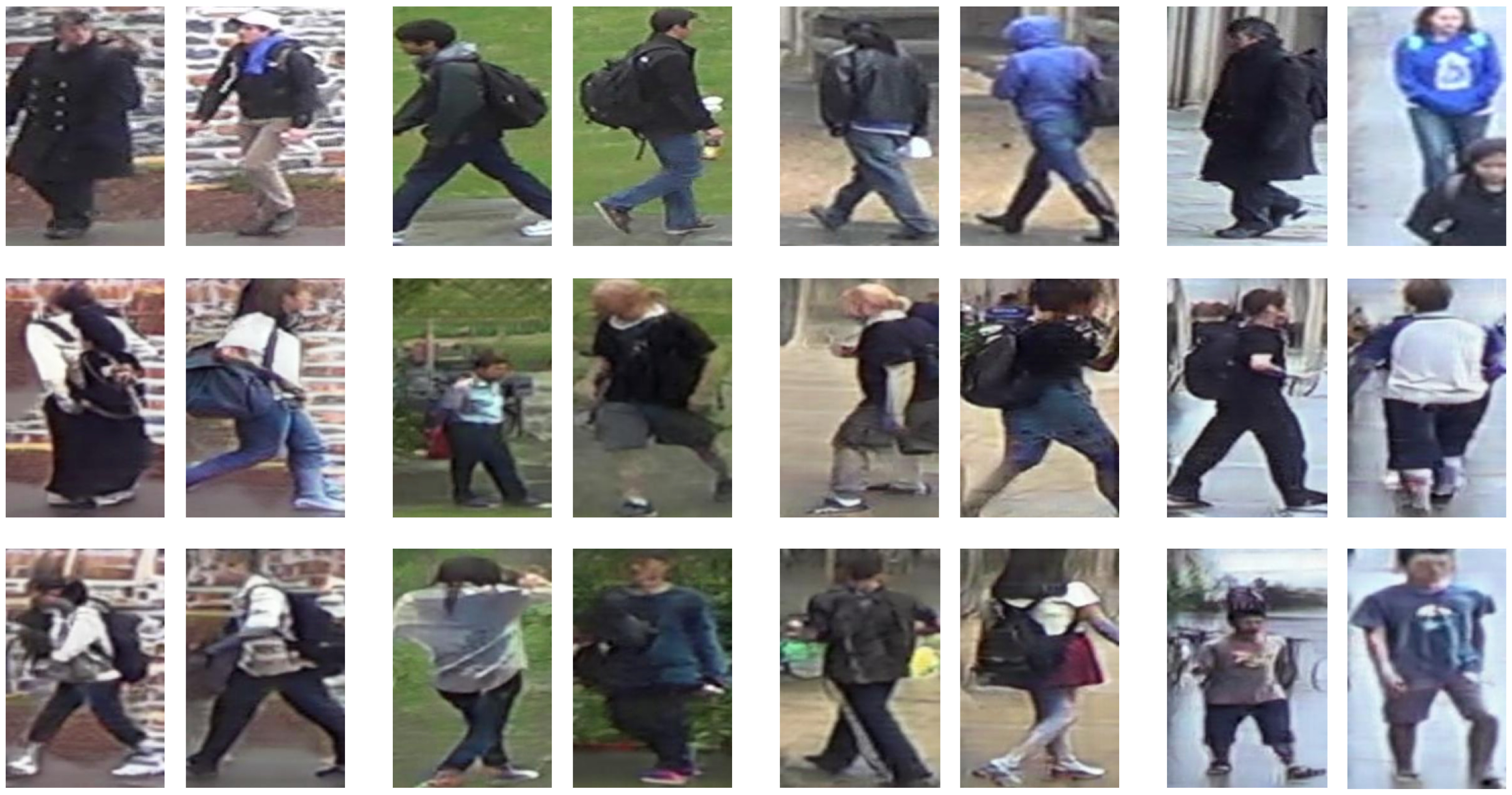}
\vspace{-2mm}
\end{center}
\caption{Illustration of the transferred person images to \emph{Duke}. The images in first row are from \emph{Duke}. The images in second and third rows are transferred images from \emph{Market} to \emph{Duke}. Obviously, those images have the similar styles, \emph{ e.g.}, similar backgrounds and lightings, \emph{etc.}}
\label{fig:fig7}
\end{figure}

The large person ReID dataset commonly contains a large number of cameras, making it expensive to perform person transfer to each individual camera. Therefore, different from the experimental settings in Sec.~\ref{sec:BigToSmall}, we do not distinguish different cameras and directly transfer person images to the target dataset with one PTGAN. Obviously, this is not an optimal solution for person transfer. Our experimental results are summarized in Fig.~\ref{fig:PTGAN_comp}.
It is obvious that GoogLeNet trained on transferred datasets works better than the one trained on the original training sets. Sample transferred images are presented in Fig.~\ref{fig:fig7}. It is obvious that, although we use a simple transfer strategy, PTGAN still generates high quality images. Possible better solutions for person transfer to large datasets will be discussed as our future work in Sec.~\ref{sec:discussion}.

\subsection{Performance of Person Transfer on \textbf{\emph{MSMT17}}}
\label{sec:pku}
We further test PTGAN on \emph{MSMT17}. We use the same strategy in Sec.~\ref{sec:BigToBig} to conduct person transfer. As shown in Table~\ref{tab:tab5}, the domain gaps between \emph{MSMT17} and the other three datasets are effectively narrowed-down by PTGAN. For instance, the Rank-1 accuracy is improved by $4.7\%$, $6.8\%$, and $3.7\%$ after performing person transfer from \emph{Duke}, \emph{Market}, and \emph{CUHK03}, respectively.

In real scenarios, the testing set is commonly large and has limited number of labeled training data. We hence test the validity of person transfer in such case. We first show the person ReID performance using different portions of training data on \emph{MSMT17} in Table~\ref{tab:tab6}. From the comparison between Table~\ref{tab:tab5} and Table~\ref{tab:tab6}, it can be observed that $10\%$ of \emph{MSMT17} training set gets similar performance with the transferred training set from \emph{Duke}, \emph{e.g.}, both achieve the Rank-1 accuracy of about 11.5\%$\sim$11.8\%. Therefore, 16,522 transferred images from \emph{Duke} achieves similar performance with 2,602 annotated images on \emph{MSMT17}. We can roughly estimate that 6.3 transferred images are equivalent to 1 annotated image. This thus effectively relieves the cost of data annotation on new datasets. The transferred data is then combined with the training set on \emph{MSMT17}. As shown in Table~\ref{tab:tab6}, the Rank-1 accuracy is constantly improved by $1.9\%$, $5.1\%$, and $2.4\%$, respectively by combining the transferred data from \emph{Duke}, \emph{Market}, and \emph{CUHK03}, respectively.

\begin{table}
\footnotesize
\caption{The performance of GoogLeNet tested on \emph{MSMT17}. The subscript ${MS}$ denotes the transferred target dataset \emph{MSMT17}.}\label{tab:tab5}

\begin{center}
\vspace{-5mm}
\resizebox{0.48\textwidth}{!}{%
\begin{tabular}{c|cc|cc|ccc}
\hline
  &\emph{Duke}&\emph{Duke}$^{*}_{MS}$  &\emph{Market} &\emph{Market}$^{*}_{MS}$ &\emph{CUHK03} &\emph{CUHK03}$^{*}_{MS}$\\
\hline
\hline
R-1 &7.1 &\textbf{11.8}       &3.4       &\textbf{10.2}   &2.8   &\textbf{6.5} \\
R-10 &17.4 &\textbf{27.4}   &10.0     &\textbf{24.4}   &8.6   &\textbf{17.2} \\
mAP &1.9 &\textbf{3.3}       &1.0        &\textbf{2.9}     &0.7   &\textbf{1.7}  \\
\hline
\end{tabular}}%
\end{center}
\end{table}

\begin{table}
\footnotesize
\begin{center}
\caption{The performance of GoogLeNet for weakly supervised learning on \emph{MSMT17}.}\label{tab:tab6}
\begin{tabular}{c|c|c|c}
\hline
{Training Set}       &R-1&R-10  &mAP\\
\hline
\hline
\emph{MSMT} (1\%) &0.9&3.6 &0.2\\
\emph{MSMT} (2.5\%) &2.0&7.4 &0.5\\
\emph{MSMT} (5\%) &6.3&18.1 &1.9\\
\emph{MSMT} (10\%) &11.5&26.9 &3.7\\
\hline
\emph{Duke} + \emph{MSMT17} (10\%)     &16.1&33.1 &5.5 \\
\emph{Duke}$^{*}_{MS}$ + \emph{MSMT17} (10\%)       &\textbf{18.0}&\textbf{36.4} &\textbf{6.2} \\
\hline
\emph{Market} + \emph{MSMT17} (10\%)     &12.6&28.5 &4.4 \\
\emph{Market}$^{*}_{MS}$ + \emph{MSMT17} (10\%)       &\textbf{17.7}&\textbf{35.9} &\textbf{6.0} \\
\hline

\emph{CUHK03} + \emph{MSMT17} (10\%)     &11.9&28.3 &4.1 \\
\emph{CUHK03}$^{*}_{MS}$ + \emph{MSMT17} (10\%)       &\textbf{14.3}&\textbf{31.7} &\textbf{4.6} \\
\hline
\end{tabular}
\end{center}
\vspace{-3mm}
\end{table}

\section{Conclusions and Discussions}\label{sec:discussion}
This paper contributes a large-scale \emph{MSMT17} dataset. \emph{MSMT17} presents substantially variants on lightings, scenes, backgrounds, human poses, \emph{etc.}, and is currently the largest person ReID dataset. Compared with existing datasets, \emph{MSMT17} defines a more realistic and challenging person ReID task.

PTGAN is proposed as an original work on person transfer to bridge the domain gap among datasets. Extensive experiments show PTGAN effectively reduces the domain gap. Different cameras may present different styles, making it difficult to perform multiple style transfer with one mapping function. Therefore, the person transfer strategy in Sec.~\ref{sec:BigToBig} and Sec.~\ref{sec:pku} is not yet optimal. This also explains why PTGAN learned on each individual target camera performs better in Sec.~\ref{sec:BigToSmall}. A better strategy is to consider the style differences among cameras to get more stable mapping functions. Our future work would continue to study more effective and efficient person transfer strategies for large datasets.

{\small
\bibliographystyle{ieee}
\bibliography{egbib}

\begin{thebibliography}{10}\itemsep=-1pt

\bibitem{chen2017beyond}
W.~Chen, X.~Chen, J.~Zhang, and K.~Huang.
\newblock Beyond triplet loss: a deep quadruplet network for person
  re-identification.
\newblock In {\em CVPR}, 2017.

\bibitem{cheng2016person}
D.~Cheng, Y.~Gong, S.~Zhou, J.~Wang, and N.~Zheng.
\newblock Person re-identification by multi-channel parts-based cnn with
  improved triplet loss function.
\newblock In {\em CVPR}, 2016.

\bibitem{cheng2011custom}
D.~S. Cheng, M.~Cristani, M.~Stoppa, L.~Bazzani, and V.~Murino.
\newblock Custom pictorial structures for re-identification.
\newblock In {\em BMVC}, 2011.

\bibitem{felzenszwalb2010object}
P.~F. Felzenszwalb, R.~B. Girshick, D.~McAllester, and D.~Ramanan.
\newblock Object detection with discriminatively trained part-based models.
\newblock {\em IEEE transactions on pattern analysis and machine intelligence},
  32(9):1627--1645, 2010.

\bibitem{gatys2016image}
L.~A. Gatys, A.~S. Ecker, and M.~Bethge.
\newblock Image style transfer using convolutional neural networks.
\newblock In {\em CVPR}, 2016.

\bibitem{geng2016deep}
M.~Geng, Y.~Wang, T.~Xiang, and Y.~Tian.
\newblock Deep transfer learning for person re-identification.
\newblock {\em arXiv preprint arXiv:1611.05244}, 2016.

\bibitem{goodfellow2014generative}
I.~Goodfellow, J.~Pouget-Abadie, M.~Mirza, B.~Xu, D.~Warde-Farley, S.~Ozair,
  A.~Courville, and Y.~Bengio.
\newblock Generative adversarial nets.
\newblock In {\em NIPS}, 2014.

\bibitem{VIPeR}
D.~Gray and H.~Tao.
\newblock Viewpoint invariant pedestrian recognition with an ensemble of
  localized features.
\newblock In {\em ECCV}, 2008.

\bibitem{hermans2017defense}
A.~Hermans, L.~Beyer, and B.~Leibe.
\newblock In defense of the triplet loss for person re-identification.
\newblock {\em arXiv preprint arXiv:1703.07737}, 2017.

\bibitem{hirzer2011person}
M.~Hirzer, C.~Beleznai, P.~M. Roth, and H.~Bischof.
\newblock Person re-identification by descriptive and discriminative
  classification.
\newblock In {\em SCIA}, 2011.

\bibitem{insafutdinov2016deepercut}
E.~Insafutdinov, L.~Pishchulin, B.~Andres, M.~Andriluka, and B.~Schiele.
\newblock Deepercut: A deeper, stronger, and faster multi-person pose
  estimation model.
\newblock In {\em ECCV}, 2016.

\bibitem{isola2016image}
P.~Isola, J.-Y. Zhu, T.~Zhou, and A.~A. Efros.
\newblock Image-to-image translation with conditional adversarial networks.
\newblock {\em arXiv preprint arXiv:1611.07004}, 2016.

\bibitem{jaderberg2015spatial}
M.~Jaderberg, K.~Simonyan, A.~Zisserman, et~al.
\newblock Spatial transformer networks.
\newblock In {\em NIPS}, 2015.

\bibitem{kim2017learning}
T.~Kim, M.~Cha, H.~Kim, J.~Lee, and J.~Kim.
\newblock Learning to discover cross-domain relations with generative
  adversarial networks.
\newblock {\em arXiv preprint arXiv:1703.05192}, 2017.

\bibitem{kinga2015method}
D.~Kinga and J.~B. Adam.
\newblock A method for stochastic optimization.
\newblock In {\em ICLR}, 2015.

\bibitem{ledig2016photo}
C.~Ledig, L.~Theis, F.~Husz{\'a}r, J.~Caballero, A.~Cunningham, A.~Acosta,
  A.~Aitken, A.~Tejani, J.~Totz, Z.~Wang, et~al.
\newblock Photo-realistic single image super-resolution using a generative
  adversarial network.
\newblock {\em arXiv preprint arXiv:1609.04802}, 2016.

\bibitem{li2016precomputed}
C.~Li and M.~Wand.
\newblock Precomputed real-time texture synthesis with markovian generative
  adversarial networks.
\newblock In {\em ECCV}, 2016.

\bibitem{li2017learning}
D.~Li, X.~Chen, Z.~Zhang, and K.~Huang.
\newblock Learning deep context-aware features over body and latent parts for
  person re-identification.
\newblock In {\em CVPR}, 2017.

\bibitem{cuhk01}
W.~Li, R.~Zhao, and X.~Wang.
\newblock Human reidentification with transferred metric learning.
\newblock In {\em ACCV}, 2012.

\bibitem{li2014deepreid}
W.~Li, R.~Zhao, T.~Xiao, and X.~Wang.
\newblock Deepreid: Deep filter pairing neural network for person
  re-identification.
\newblock In {\em CVPR}, 2014.

\bibitem{liao2015person}
S.~Liao, Y.~Hu, X.~Zhu, and S.~Z. Li.
\newblock Person re-identification by local maximal occurrence representation
  and metric learning.
\newblock In {\em CVPR}, 2015.

\bibitem{liu2016coupled}
M.-Y. Liu and O.~Tuzel.
\newblock Coupled generative adversarial networks.
\newblock In {\em NIPS}, 2016.

\bibitem{liu2017hydraplus}
X.~Liu, H.~Zhao, M.~Tian, L.~Sheng, J.~Shao, S.~Yi, J.~Yan, and X.~Wang.
\newblock Hydraplus-net: Attentive deep features for pedestrian analysis.
\newblock In {\em ICCV}, 2017.

\bibitem{mirza2014conditional}
M.~Mirza and S.~Osindero.
\newblock Conditional generative adversarial nets.
\newblock {\em arXiv preprint arXiv:1411.1784}, 2014.

\bibitem{reed2016generative}
S.~Reed, Z.~Akata, X.~Yan, L.~Logeswaran, B.~Schiele, and H.~Lee.
\newblock Generative adversarial text to image synthesis.
\newblock {\em arXiv preprint arXiv:1605.05396}, 2016.

\bibitem{ren2015faster}
S.~Ren, K.~He, R.~Girshick, and J.~Sun.
\newblock Faster r-cnn: Towards real-time object detection with region proposal
  networks.
\newblock In {\em NIPS}, 2015.

\bibitem{ristani2016MTMC}
E.~Ristani, F.~Solera, R.~Zou, R.~Cucchiara, and C.~Tomasi.
\newblock Performance measures and a data set for multi-target, multi-camera
  tracking.
\newblock In {\em ECCV workshop}, 2016.

\bibitem{su2017pose}
C.~Su, J.~Li, S.~Zhang, J.~Xing, W.~Gao, and Q.~Tian.
\newblock Pose-driven deep convolutional model for person re-identification.
\newblock In {\em ICCV}, 2017.

\bibitem{szegedy2015going}
C.~Szegedy, W.~Liu, Y.~Jia, P.~Sermanet, S.~Reed, D.~Anguelov, D.~Erhan,
  V.~Vanhoucke, and A.~Rabinovich.
\newblock Going deeper with convolutions.
\newblock In {\em CVPR}, 2015.

\bibitem{wang2016generative}
X.~Wang and A.~Gupta.
\newblock Generative image modeling using style and structure adversarial
  networks.
\newblock In {\em ECCV}, 2016.

\bibitem{wei2017glad}
L.~Wei, S.~Zhang, H.~Yao, W.~Gao, and Q.~Tian.
\newblock Glad: Global-local-alignment descriptor for pedestrian retrieval.
\newblock In {\em ACM MM}, 2017.

\bibitem{wu2016enhanced}
S.~Wu, Y.-C. Chen, X.~Li, A.-C. Wu, J.-J. You, and W.-S. Zheng.
\newblock An enhanced deep feature representation for person re-identification.
\newblock In {\em WACV}, 2016.

\bibitem{xiao2016learning}
T.~Xiao, H.~Li, W.~Ouyang, and X.~Wang.
\newblock Learning deep feature representations with domain guided dropout for
  person re-identification.
\newblock In {\em CVPR}, 2016.

\bibitem{yan2016attribute2image}
X.~Yan, J.~Yang, K.~Sohn, and H.~Lee.
\newblock Attribute2image: Conditional image generation from visual attributes.
\newblock In {\em ECCV}, 2016.

\bibitem{yi2017dualgan}
Z.~Yi, H.~Zhang, P.~T. Gong, et~al.
\newblock Dualgan: Unsupervised dual learning for image-to-image translation.
\newblock {\em arXiv preprint arXiv:1704.02510}, 2017.

\bibitem{yoo2016pixel}
D.~Yoo, N.~Kim, S.~Park, A.~S. Paek, and I.~S. Kweon.
\newblock Pixel-level domain transfer.
\newblock In {\em ECCV}, 2016.

\bibitem{zhao2017pspnet}
H.~Zhao, J.~Shi, X.~Qi, X.~Wang, and J.~Jia.
\newblock Pyramid scene parsing network.
\newblock In {\em CVPR}, 2017.

\bibitem{zhao2017spindle}
H.~Zhao, M.~Tian, S.~Sun, J.~Shao, J.~Yan, S.~Yi, X.~Wang, and X.~Tang.
\newblock Spindle net: Person re-identification with human body region guided
  feature decomposition and fusion.
\newblock In {\em CVPR}, 2017.

\bibitem{zheng2015scalable}
L.~Zheng, L.~Shen, L.~Tian, S.~Wang, J.~Wang, and Q.~Tian.
\newblock Scalable person re-identification: A benchmark.
\newblock In {\em ICCV}, 2015.

\bibitem{zheng2016discriminatively}
Z.~Zheng, L.~Zheng, and Y.~Yang.
\newblock A discriminatively learned cnn embedding for person
  re-identification.
\newblock {\em arXiv preprint arXiv:1611.05666}, 2016.

\bibitem{zheng2017unlabeled}
Z.~Zheng, L.~Zheng, and Y.~Yang.
\newblock Unlabeled samples generated by gan improve the person
  re-identification baseline in vitro.
\newblock In {\em ICCV}, 2017.

\bibitem{zhu2017unpaired}
J.-Y. Zhu, T.~Park, P.~Isola, and A.~A. Efros.
\newblock Unpaired image-to-image translation using cycle-consistent
  adversarial networks.
\newblock In {\em ICCV}, 2017.

\end{thebibliography}
}

\end{document}